%% file: acl2023.tex
\pdfoutput=1
\documentclass[11pt]{article}

\usepackage[]{ACL2023}

\usepackage{times}
\usepackage{latexsym}

\usepackage[T1]{fontenc}
\usepackage{pifont}
\usepackage[utf8]{inputenc}

\usepackage{microtype}

\usepackage{inconsolata}
\usepackage{amssymb}
\usepackage{multirow}
\usepackage{tabularx}
\usepackage{caption}
\usepackage{graphicx}
\usepackage{subcaption}
\usepackage{booktabs} 
\newcommand{\oursystem}{$\mathsf{EcoRank}$}

\newcommand{\cmark}{\ding{51}}

\title{EcoRank: Budget-Constrained Text Re-ranking Using Large Language Models}

\author{Muhammad Shihab Rashid, Jannat Ara Meem, Yue Dong, Vagelis Hristidis \\
    University of California, Riverside \\
    \texttt{\{mrash013, jmeem001, yue.dong\}@ucr.edu, vagelis@cs.ucr.edu}}

\begin{document}
\maketitle
\begin{abstract}
\input{Sections/a_abstract}
\end{abstract}

\input{Sections/b_introduction}

\input{Sections/c_problem_definition}
\input{Sections/d_ranking_algo}
\input{Sections/e_evaluation}

\input{Sections/f_related_work}
\input{Sections/h_conclusion}
\newpage
\section*{Limitations}
\input{Sections/i_limitation}
\section*{Ethics Statement}
Although our work involves the usage of LLMs, which are known to hallucinate, it does not create any ethical concerns. We use LLMs to re-rank an already ranked list hence possible hallucinations do not provide any harmful effect. We adhere to the Code of Ethics with our work. No personal or restricted data were collected from any source or subject.

%\section*{Acknowledgements}

\bibliography{custom}
\bibliographystyle{acl_natbib}

\appendix

\input{Sections/g_appendix}

\end{document}

%% file: Sections/a_abstract.tex
Large Language Models (LLMs) have achieved state-of-the-art performance in text re-ranking. This process includes queries and candidate passages in the prompts, utilizing pointwise, listwise, and pairwise prompting strategies. A limitation of these ranking strategies with LLMs is their cost: the process can become expensive due to API charges, which are based on the number of input and output tokens.
%For business which deal with thousands of queries daily, the cost may exceed the \textit{budget}, which makes it their biggest constraint. 
% Recent work has studied how to reduce the cost spent on LLMs using LLM cascading. However, 
% %they either require a fine-tuned model with training data or 
% these methods are not applicable to text re-ranking tasks. 
We study how to maximize the re-ranking performance given a \textit{budget}, by navigating the vast search spaces of prompt choices, LLM APIs, and budget splits. We propose a suite of budget-constrained methods to perform text re-ranking using a set of LLM APIs. Our most efficient method, called {\oursystem}, is a two-layered pipeline that jointly optimizes decisions regarding budget allocation across prompt strategies and LLM APIs. Our experimental results on four popular QA and passage reranking datasets show that EcoRank outperforms other budget-aware supervised and unsupervised baselines.
%We also provide analysis on different budget categories and the performance of various methods.

%% file: Sections/b_introduction.tex
\section{Introduction}\label{sec:intro}
Text re-ranking focuses on ranking $N$ source documents given a specific query and is crucial for providing the relevant retrieved context to downstream tasks. It serves either as a standalone task or as an intermediate step for question answering tasks~\citep{rashid2024normy} in a retrieval augmented setting, where the answer is generated from the top $k$ relevant passages. Traditional ranking methods includes BM25~\citep{bm25} and neural methods like DPR~\citep{karpukhin2020dense}, Contriever~\citep{izacard2021contriever} etc. Recently, large language models (LLMs) such as GPT-4~\citep{openai2023gpt} 
have demonstrated dominant performance in text re-ranking~\citep{sun2023chatgpt}. 

However, utilizing LLMs often comes at a cost: the process can be quite expensive, as closed-source LLMs charge based on the number of tokens. This expense escalates in text re-ranking due to the need to input substantial text, proportional to the number of passages to re-rank. For example, re-ranking 500 passages for a single query, with each passage having an average length of 100 tokens, currently costs at least 5 USD when using GPT-4~\citep{openaiprice}. This cost becomes intractable when businesses need to handle thousands of queries daily, making \emph{budget} the biggest constraint.

Although there are many alternatives such as TextSynth~\citep{textsynth}, AI21~\citep{ai21}, Cohere~\citep{cohere}, Replicate~\citep{replicate}, etc., that offer LLM API services at lower costs, utilizing these commercial APIs may still not be sustainable with high volumes of queries. This motivates us to investigate the trade-off between cost and performance for text re-ranking. Our aim is to propose a budget-aware solution for text re-ranking that maximizes performance within the constraints of a given budget. With this goal in mind, we approach budget-constrained text re-ranking using LLMs as a constrained optimization problem. Here, we explore various re-ranking methods with different properties and optimize for the best strategy for budget-aware text re-ranking. 
\begin{figure*}[t]
\centering
  \includegraphics[width=0.72\textwidth]{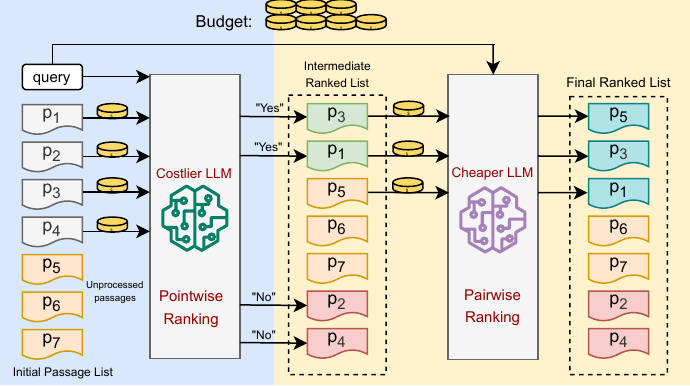}
  \caption{An overview of {\oursystem} with an example of 7 passages. A fraction of budget is spent on 4 passages for pointwise prompt with a costlier LLM and an intermediate ranked list is generated with unprocessed passages in the middle. Then, using the rest of the budget we call the cheaper LLM to do pairwise comparisons and create the final ranked list.}
  \label{fig:overview}
\end{figure*}

Our work contributes to the first efforts in budget-aware modeling utilizing LLMs for text re-ranking, to the best of our knowledge. Recent work on cost-aware applications of LLMs with \textit{LLM Cascading} \citep{chen2023frugalgpt, yue2023large, vsakota2023fly} is not applicable to text re-ranking. They either focus primarily on QA or reasoning tasks \citep{yue2023large}, or require a fine-tuned model (with training data) to assess the generation quality of LLMs \citep{chen2023frugalgpt, vsakota2023fly}.
%\yue{check the claim, not sure whether this is the reason that they cannot be used for re-ranking} \shihab{Two papers need ML training data to score the generation of LLM, the other paper do not need training but they are not applicable to text re-ranking as they use answer sampling, which is not possible in our case}. 
On the other hand, works focusing on using LLMs for text re-ranking primarily aim at performance improvement without considering budgets. The three most common approaches exhibit increasing costs as the number of tokens inputted into the LLMs increases: 1) Pointwise prompts \citep{sachan2022improving}, which input a single passage per request and output calibrated prediction probabilities before sorting; 2) Listwise prompts \citep{sun2023chatgpt, ma2023zero}, which input multiple passages per request as lists and ask LLMs to output the lists in order; 3) Pairwise prompts \citep{qin2023large}, which input pairs of passages per query per request and use a sliding window to sort the top-k passages. Approximately, the most expensive approach, namely pairwise prompts, can cost about $2 \cdot k$ times more than pointwise prompts where $k$ is the size of the sliding window. 

In this paper, we propose a suite of budget-constrained methods to perform text reranking using a set of LLM APIs.  Our most efficient method, which we refer as {\oursystem}, is a budget-constrained LLM-based text re-ranking pipeline, that jointly optimizes several objectives: 1) which prompt designs to deploy, 2) which LLM APIs to call, and 3) how to split budget between multiple prompts and LLMs.
Optimizing all these decisions jointly is challenging for the following reasons. 
(a) The provided budget may not  be enough to input all input texts once to the LLM API. The developed method must be able to optimize for the top-few (e.g. top-1) text.
(b) Different LLM APIs may have different strengths and limitations.
% (c) We do not have any training data on how to best split the budget.
(c) There is an exponential number of combinations of prompt designs and API selections.

% An interesting observation is that {\oursystem} achieves the best performance even if there was no budget constraint, which proves the efficacy of our pipeline. 

Addressing these questions, we first consider various text re-ranking prompts tailored to accomplish the re-ranking task within a budget. Next, we explore different LLM APIs and their associated costs for implementing these prompts. We then introduce a novel two-layer approach, {\oursystem} (depicted in Figure~\ref{fig:overview}), which begins by re-ranking initially ranked passages (e.g., those ranked using BM25) with \textit{pointwise relevance filtering} on a \textit{high-accuracy} (and consequently \textit{expensive}) LLM API, utilizing a fraction of our budget. This initial re-ranking demotes irrelevant passages, allowing us to allocate the remaining budget to re-rank relatively relevant passages. In the second layer, we use a \textit{less accurate} (and thus \textit{cheaper}) LLM API, applying the remaining budget to further re-rank the passages using \textit{pairwise ranking prompting}.
%This ensures the relative ordering among the relevant passages. {\oursystem} is able to overcome the weaknesses of each individual prompt design and at the same time optimizes the cost by intelligently picking which passages to spend budget on. It is completely zero-shot, requires no training data and is able to produce state-of-the-art ranking performance.

We evaluate various single or hybrid (combining more than one) prompt designs for the text re-ranking problem, for various budgets and APIs on four popular datasets: Natural Questions (NQ)~\citep{kwiatkowski2019natural}, Web Questions (WQ)~\citep{berant2013semantic}, TREC~\citep{craswell2020overview} DL19, and DL20. Our most efficient method {\oursystem} achieves a gain of 14\% on MRR and R@1 ranking accuracy than baselines.

Our contributions are summarized as follows:
\begin{itemize}
    \item We introduce the problem of budget-constrained text re-ranking that considers the cost of various LLM APIs.
    \item We propose and compare various ranking prompt designs and API choices in a budget-constrained scenario for text re-ranking.
    \item We further propose a novel two-layer cascading pipeline {\oursystem} which optimizes the budget usage.
    %by intelligently making a decision on the input to spend budget on which achieves state-of-the-art performance.
    \item We extensively evaluate and compare various prompt designs and API choices on four datasets. We make our code available to the research community.~\footnote{\url{https://github.com/shihabrashid-ucr/EcoRank}}
\end{itemize}

%The rest of the paper is organized as follows. We define the scope and problem statement in Section~\ref{sec:problem}. We explain in details all the ranking prompt designs in Section~\ref{sec:ranking_algo} and most efficient approach {\oursystem} in Section~\ref{sec:our_system}. We present the results of our experimental evaluation in Section~\ref{sec:evaluation}. We discuss the related work in Section~\ref{sec:related} and finally conclude in Section~\ref{sec:conclusion}.

%% file: Sections/c_problem_definition.tex
\section{Problem Definition}\label{sec:problem}
% The input is a pre-ranked list of $N$ texts (i.e. passages) based on query $q$, a budget $\beta$, and a set of LLM APIs. The output is a re-ranked list of the $N$ texts.

We focus on optimizing passage re-ranking under budget constraints, ranking top-$k$ passages from a pre-ranked list. These lists are often obtained from a retriever in response to a natural language query. Given a budget $\beta$, a query $q$, and a list of pre-ranked $N$ passages $p_0 \cdots p_N$ with respect to $q$, the task is to re-rank the passages using available LLM APIs and retrieve top-$k$ passages.  For instance, in question answering tasks, BM25 is commonly used to produce the initial ranking, and then the main focus is typically on the top re-ranked passage (i.e. $k=1$). 
%There are several supervised  or unsupervised  transformer-based algorithms for text re-ranking that do not use any LLM~\citep{nogueira2020document, asai2022task}. However, those are out of scope for this paper. 
The following designs are critical for budget-aware text re-ranking.

\paragraph{API choice.} Assume there are \textit{M} different LLM APIs available, denoted as $\mathcal{L}_1\cdots\mathcal{L}_M$. Each API $\mathcal{L}$ takes a prompt $\rho$ and generates an output $\Theta$. Associated with calling each API is a cost $\mathcal{C}$, defined in Equation~\ref{eq:llm_cost}: %\yue{why do you use these notations, is it better to use x as input, y as output?}\shihab{no particular reason, we use x and y as other parameters, $rho$ looks like p, which is prompt, p may be confusing. Same for $Theta$, o = output}
\begin{equation}
\label{eq:llm_cost}
    \mathcal{C} = c_p \cdot len(\rho) + c_o \cdot len(\Theta) + c_f
\end{equation}
where $c_p$  represents the cost per input or prompt token, $c_o$ is the cost determined by the number of tokens generated by $\mathcal{L}$, and occasionally, a fixed API call cost $c_f$ applies. 
%\yue{similarly, why $c_p$ and $c_o$, is it better to be $c_i$ and $c_o$?}\shihab{we can do $c_i$ where i=input, i wrote $c_p$, p = prompt.}

\paragraph{Choice of ranking prompt.} The other key design choice is the set of prompts $\rho \in T$  for a text re-ranking task. Specifically, given the passages $p_0\cdots p_N$ and query $q$, we need to generate a sequence of prompts, where each prompt includes the query and one or more of the input passages. For example, a prompt can ask if a passage is relevant or ask to compare two passages. As input tokens determines the cost to generate an output, choosing the right prompt(s) is very crucial.

\paragraph{Split of budget.} Our experimental setup also accounts for scenarios where multiple prompts or rounds of iterations are required. In such cases, there is an additional factor in budget considerations: the budget $\beta$ can be divided and allocated across multiple prompts $\rho$, $\beta \triangleq \beta \cdot x + \cdots +\beta \cdot y$ where the coefficients $x+ \cdots + y = 1$\\

\paragraph{Budget-aware optimization.} 
Given the problem setups described above, the task of budget-aware text re-ranking essentially becomes an optimization task. The objective is to maximize the re-ranking performance, denoted as \( \mathbb{E} \), subject to a given budget \( \beta \). This optimization occurs within the search spaces of prompts, APIs, and budget allocations, $\mathbb{E}_{q, p}[c (\mathcal{L}, \rho)] \leq \beta$, where $c(\mathcal{L}, \rho)$ is the associated cost for processing query $q$ with prompt $\rho$ and LLM $\mathcal{L}$. Given the vast range of available APIs, prompts, and permutations of budget splits, this optimization presents non-trivial and unique challenges.

%% file: Sections/d_ranking_algo.tex
\section{Budget-Aware Ranking Prompt Designs} \label{sec:ranking_algo}
\begin{figure}[!ht]
\centering
  \includegraphics[width=0.79\linewidth]{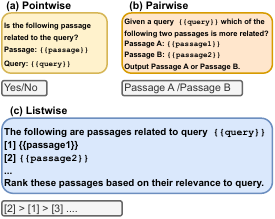}
  \caption{Different prompt strategies for text re-ranking.}
  \label{fig:prompts}
\end{figure}
We build on previous works on ranking prompt designs. Our key contribution is making these designs budget-constrained.
%For text re-ranking with LLMs, there can be three types of strategies for prompt designs. 
The different designs are depicted in Figure~\ref{fig:prompts}.

\subsection{Pointwise Methods}
Pointwise approaches process the passages one by one along with the query as a prompt. We define three types of prompts within pointwise methods.

\paragraph{Query generation.} \citet{sachan2022improving} proposed an unsupervised pipeline \textit{UPR} to re-rank passages by asking the LLM to generate a query $q'$ given each passage $p_i$ from the initial ranked list of passages. 
%They extract the calibrated prediction probabilities from the APIs which is equivalent to the confidence of outputs of the LLM models. Based on the log-likelihood of the scores they sort the passages. 
Their approach is not applicable to generation-only LLMs like GPT-3 or GPT-4 which do not score the outputs.
For budget constrained scenarios, we adapt their approach to generation only LLMs by asking the LLM to generate a query given a passage and measure the token-level F1 score between the newly generated query $q'$ and original query $q$ and sort the passages based on this score. 
%Higher score means high matching between the queries. The hypothesis being that a passage generating a query similar to the original one is highly relevant. 
We start from the top of the list and go down until we exhaust the budget, and keep the rest in their original ranking positions. We call this approach \texttt{B-UPR}.

\paragraph{Binary Classification.} We propose another type of budget-aware pointwise prompt design inspired from~\citet{liang2022holistic} where we ask the LLM to predict \textit{Yes} or \textit{No} given each passage from the initial list whether it is relevant to the query $q$. 
%The prompt is shown in Figure~\ref{fig:prompts}. 
We ask the LLM to output only "Yes" or "No" which restricts the number of output tokens to be 1. 
%Multiple passages may get the output of "Yes" as the passages are initially retrieved by a retriever like BM25. For a passage, if the output is not parseable or we exhaust the budget, we fall back to the initial ranking. This results in a list of re-ranked passages where all the "Yes" passages are grouped together at the top sorted by the initial ordering, unprocessed passages at the middle, and "No" passages at the bottom.
This method can be categorized as a \textit{coarse-grained} strategy where we group the passages based on relevance but each individual passage ranking relies on the initial score. 
%However, more passages can be processed with LLMs as it requires less number of tokens as input and output than other methods.

\paragraph{Likert Classification.} Instead of classifying each passage in a binary fashion, we also introduce a design to categorically classify each passage into a 3-point Likert scale, inspired from~\citet{zhuang2023beyond}, where we ask the LLM to classify a passage into either of the following three groups: \textit{Very related, Somewhat related} or \textit{Unrelated}. 
%This is an attempt to fine-grain the passages a little bit more with the same number of tokens. 
%However, it may be difficult for an LLM to differentiate between \textit{Very related} and \textit{Somewhat related}. Similar to previous method, we group the \textit{Very related} passages at the top, then \textit{Somewhat related}, then unprocessed passages, and \textit{Unrelated} passages at the bottom.
For more details on how we rank the passages using pointwise, please refer to~\ref{app:rank_pointwise}.
\subsection{Listwise Methods}
In this strategy, passages $p_0 \cdots p_N$ are put through the LLMs with identifiers such as ([1], [2], etc.) as a list along with the query $q$. The LLM is then asked to give the relative ordering of the passages as an output (i.e. [2] > [1] $\cdots$). 
%As LLMs have a token limit usually of 4096 tokens or less, it may not be possible to include all $N$ passages in a single prompt. 
Recent works like RankGPT~\citep{sun2023chatgpt}, LRL~\citep{ma2023zero} introduce a sliding window strategy to combat token limitation challenge in LLMs, where a sliding window of size $w$ with a step of $s$ is used. 
%In this way, this strategy is able to promote the $w$ most relevant passages to the top of the list in one round. 
%However, when there is a budget, it may not be possible to include $w$ passages in a single prompt. Therefore, 
For a budget-constrained scenario, we approximate the number of passages that can be given as input to the prompt. %using the token length of the passages, query, and possible output given the budget. 
%The rest of the passages not processed by LLMs are ranked according to initial score. 
We call this \texttt{B-RankGPT}.

However, as the prompt of listwise method is rather complicated (LLMs have to understand the ordering of multiple passages and have to give an output in a structured format), most LLM APIs face issues in giving the correct output. \citet{qin2023large} show that medium-sized LLM APIs like FLAN-T5-XL~\citep{chung2022scaling} are inconsistent and not able to understand the prompt correctly and provide irrelevant results. 
%Sometimes, the LLMs only output a partial list or do not output the identifier ([1], [2] etc.) and produces text which is not parseable. 
Only big-sized commercial LLMs like GPT-4 and GPT-3.5 are able to utilize this approach correctly. 
%However, they are much more expensive than medium sized LLM APIs and may not work the best when working within a budget. 
Further, it is highly sensitive to input ordering, meaning the output depends heavily on the order of the passages in the prompt. 
%which is not ideal as it reinforces the initial ordering.
% \begin{figure*}[t]
% \centering
%   \includegraphics[width=0.75\textwidth]{Figures/abc_overview_v3.pdf}
%   \caption{An overview of {\oursystem} with an example of 7 passages. A fraction of budget is spent on 4 passages for pointwise prompt with a costlier LLM and an intermediate ranked list is generated with unprocessed passages in the middle. Then, using the rest of the budget we call the cheaper LLM to do pairwise comparisons and create the final ranked list.}
%   \label{fig:overview}
% \end{figure*}
\input{Tables/llm_variants}

\subsection{Pairwise Methods}
%Pairwise methods include two passages in one prompt and compare them given a query. 
This is a \textit{fine-grained} strategy where each passage is compared with each other similar to bubble-sort and their ranking is modified. Given two passages $p_i$, $p_j$, and one query $q$, the LLM is asked to choose one passage which is more relevant to the query. This ensures the \textit{relative} ordering among the passages. \citet{qin2023large} proposed various pairwise ranking prompts (PRP). Among them, \textit{PRP-Sliding}, which does $k$ rounds of bubble-sort pass that ensures the top-$k$ ranking performs best.
%\citet{qin2023large} proposed three types of pairwise ranking prompts (PRP): \textit{1) PRP-Allpair.} All possible pairs of passages are enumerated and compared. This is similar to a full bubble-sort algorithm with $O(N^2)$ complexity, where N is the total number of passages. \textit{2) PRP-Sorting.} This is based on heapsort algorithm where the complexity is $O(NlogN)$. Each pairwise comparison is one LLM API call. \textit{3) PRP-Sliding.} PRP-Sliding does $k$ rounds of bubble-sort approach which ensures the top-$k$ ranking.

However, all the approaches, even sliding-k are quite expensive. To get top-$k$ ranking, $N\cdot k$ API calls need to be made with each call having approximately twice the number of tokens (because two passages per prompt) compared to pointwise methods, which need to make $N$ calls with each call having less tokens. In contrast, PRP methods can achieve fine-grained ranking accuracy. We adapt this strategy to a budget-constrained one by approximating the number of calls that could be made within the budget. If $\tau$ such calls can be made, we start at $l = min(k, \tau)$-th positioned passage in the initial ranked list and move the passage up the list. We continue iterating until $\tau=0$. For passages that could not be processed, we take their initial ranking position. We call this approach \texttt{B-PRP}.

\section{\oursystem}\label{sec:our_system}
In this section, we present our novel most efficient budget-aware approach shown in Figure~\ref{fig:overview}. Both listwise and pairwise methods can achieve accurate rankings but they suffer from high cost. Listwise methods only work with very expensive LLMs hence we put our focus on pairwise approach. While constrained within a budget, the number of passages that can be processed by LLMs impacts the final ranking accuracy. Cheaper LLMs are able to process more passages but they lack quality. Therefore, there are two key challenges that needs to be solved: 1) how to ensure quality 2) how to ensure quantity. With no budget constraint, pairwise methods may be an obvious choice due to their fine-grained accuracy but they have several limitations. We solve the challenges of pairwise designs in a budget-constrained scenario in a two-staged fashion as shown in Figure~\ref{fig:overview}.

\paragraph{First stage.} Only the first few passages can be compared by pairwise with a limited budget. However, as the initial ranked list is not that good, valuable tokens may be spent on irrelevant passages. Thus in the first stage, we intelligently pick the passages to do pairwise comparison on.

We split our budget $\beta$ into two fractions $x$ and $y$ and use $x$ amount to filter the passages using binary classification approach from the pointwise prompting group. We use a stronger and comparatively expensive LLM $\mathcal{L}_1$ with cost $\mathcal{C}_1$ to generate a relevance for the passages in the form of "Yes" or "No" and create an intermediate ranked list as per the binary classification strategy mentioned above. This coarse-grained strategy ensures \textit{quality} as we are using a strong LLM to put more relevant passages at the top for the next stage and push irrelevant ones to the bottom of the list.

\paragraph{Second stage.} We spend the rest $y$ amount using pairwise prompting design to compare two passages at a time. To ensure \textit{quantity}, we use a cheaper LLM API $\mathcal{L}_2$ with cost $\mathcal{C}_2$ to do the comparisons. As the passages have already been filtered by a stronger LLM, a cheaper LLM does not hinder the quality that much rather it can process $(\mathcal{C}_1 / \mathcal{C}_2)$x the number of passages than the expensive variant for a fixed budget. Further, as there are much fewer "Yes" passages compared to "No", the unprocessed passages from first stage can be processed at this stage.

While further stages can be added to the pipeline, we see diminishing results as we add more stages as discussed in Appendix~\ref{app:n_stages}. Thus we choose two.

\paragraph{Choosing the LLMs.} As seen in Table~\ref{tab:llm_variants}, LLMs' performance is not proportional to their cost. Some expensive API (i.e. GPT-curie or GPT-3.5) may be less effective in zero-shot text re-ranking tasks than comparatively cheaper LLM (i.e. T5-L or T5-XL). Further, same-priced LLMs may not give the same performance. A key optimization here is to choose the appropriate LLM which will perform the task accurately without much cost. As the T5 LLMs perform significantly better with reasonable cost, we choose Flan T5-XL as the costlier API $\mathcal{L}_1$ and Flan T5-L as the cheaper API $\mathcal{L}_2$ in {\oursystem}. There is an opportunity to automatically choose the appropriate LLM which we discuss in Appendix~\ref{app:automate}.  

\paragraph{Optimization of budget split.} Another key challenge here is the split of budget $x$ and $y$. Putting more budget in the first stage will ensure more \textit{quality} filtering whereas putting more budget on the second stage will do more pairwise but on less relevant passages. We hypothesize that both stages contribute equally to the pipeline and choose an equal split of $x$ and $y$. We experiment on various splits on two datasets and confirm our hypothesis in Section~\ref{sec:evaluation}.

%% file: Tables/llm_variants.tex
\begin{table*}[t]
\small
     \centering
    \begin{tabular}{c|c|c|c||c|c|c||c|c|c}
    \toprule
         \multirow{2}{*}{LLM}&\multirow{2}{*}{Parameters}   &\multirow{2}{*}{Cost$^\ddagger$}&\multirow{2}{*}{Provided By}  &  \multicolumn{3}{c||}{Methods (MRR)}&  \multicolumn{3}{c}{Methods (R@1)}\\
         %\cline{4-9}
         &   &&  &  \multicolumn{1}{c}{Pointwise}&  \multicolumn{1}{c}{PRP}&  \multicolumn{1}{c||}{Listwise}&  Pointwise&  PRP& Listwise\\
         \midrule
         FLAN T5-XL&  3B &1x&  Replicate&  42.08&  \textbf{46.54}&  NA&  33.4&  \textbf{39.5}& NA\\
         FLAN T5-L&  800M &$\frac{1}{3}$x&  Replicate&  37.91&  38.03&  NA&  27.4&  29.2& NA\\
         Llama2&  7B &1x&  TextSynth&  30.24&  21.18&  NA&  20.2&  10.6& NA\\
         Falcon&  7B &1x&  TextSynth&  30.8&  23.44&  NA&  20.6&  12.8& NA\\
         GPT-curie&  6.7B &5x&  OpenAI&  30.61&  19.52&  NA&  20.8&  9.40& NA\\
         GPT-3.5-turbo&  175B &10x&  OpenAI&  34.41&  43.82&  42.3&  23.9&  36.60& 34.3\\
         \bottomrule
    \end{tabular}
    \caption{Different LLMs' performance with various strategies on a subset of NQ dataset for top-20 passages. $^\ddagger$ Cost is measured as a unit here as the pricing may vary with time.}
    \label{tab:llm_variants}
    
\end{table*}

%% file: Sections/e_evaluation.tex
\section{Experimental Evaluation} \label{sec:evaluation}
%In this section we show the comparative evaluation of different approaches and give details about our implementation and datasets.
\subsection{Setup}
\paragraph{Implementation Details.} We host the LLMs offline (except for GPT3.5) and define the budget in terms of number of tokens. The GPU instance we use is g5.4xlarge. The prices of LLMs change with time frequently so we make a standard approximation of the costs of each LLM shown in Table~\ref{tab:llm_variants}. For example, T5-L is approximated as 3x cheaper than T5-XL and GPT-3.5 is 10x costlier than T5-XL based on the pricing page of different services~\cite{openaiprice, replicate, textsynth}.

 To implement the supervised models, we load the available pre-trained versions in our GPU and use them to generate embeddings of the passages and queries. We use PyGaggle~\footnote{https://github.com/castorini/pygaggle} to re-rank the passages with the loaded model.

To implement InPars, We use GPT-3.5 to generate synthetic data $d$ fully using our budget B1, B2, B3. Using B1 we could generate around 50K, using B2 10K, and using B3 5K questions. Thus we have three sets of training data $d_{b1}$, $d_{b2}$, and $d_{b3}$. We randomly sampled passages from the corpus and asked GPT-3.5 to generate a new question. These are our positive examples. To get the negative examples, we use BM25 to retrieve 5 relevant passages. The passages that are not gold passages are considered as negative examples. We trained the T5-large~\footnote{https://huggingface.co/google-t5/t5-large} models for three budget categories with 156 steps and the same training arguments as InPars. We use the corresponding trained model to generate embeddings of the passages and re-rank.
\paragraph{Datasets.} Following previous work on passage retrieval, we choose the popular benchmark datasets Natural Questions (NQ)~\citep{kwiatkowski2019natural}, Web Questions (WQ)~\citep{berant2013semantic}, TREC~\citep{craswell2020overview} DL19, and DL20. There are total of 3610 questions on NQ, 2032 on WQ, 30 on DL19, and 44 on DL20 test splits. For TREC datasets, there are multiple relevant passages per query contrary to NQ and WQ. To ensure fairness among datasets, we consider the passages with a score of 3 to be the relevant ones for TREC.
% As the evidence passage collection, we use the preprocessed English Wikipedia dump from December 2018 released by~\citet{karpukhin2020dense}. Each article is split into overlapping 100 words passages.

\input{Tables/budget_category}

\paragraph{Budget categories.} We experiment with three budget categories. Budgets are represented as token limits per question. 
%They are approximated from the pricing pages of different services. 
The number of tokens that can be processed with each LLM is mentioned in Table~\ref{tab:budget_category}. We choose a relatively higher budget B1 which can process most passages with pointwise and complete one pass with pairwise, and lower budgets B2, B3 where only some passages can be processed to show the efficacy of our solutions. For more details on the selection process, please refer to Appendix~\ref{app:selecting_budget}.
\input{Tables/main_results}

\paragraph{Baselines.} On top of the budget-constrained methods introduced above, we show comparisons with state-of-the-art supervised and unsupervised baselines. The supervised baselines are: 1) \textbf{monoBERT}~\citep{nogueira2019passage}: A cross-encoder reranker trained on BERT-large, trained on MSMARCO, 2) \textbf{monoT5}~\citep{nogueira2020document}: A sequence-to-sequence reranker based on T5, and 3) \textbf{TART}~\citep{asai2022task}: A supervised instruction-tuned passage reranker based on FLAN-T5-XL

For unsupervised approach, we consider \textbf{InPars}~\citep{bonifacio2022inpars}, where we generate synthetic training data using GPT-3 and use them to train a T5-large model. For each budget category, we spend all our budget to generate synthetic data and infer the trained model in a zero-shot approach. We also consider OpenAI \textit{text-ada-002} embedding model as a reranker baseline.

We assume zero cost for the supervised baselines and no inference cost for InPars as no paid API is used. In reality, supervised models can have some costs regarding time and computing resources. For more details, please refer to~\ref{app:supervised_cost}.
%We consider the budget-constrained methods mentioned in the previous section for evaluation. The methods are: BM25 (initial ranking); Pointwise: Binary classification, Likert classification, B-UPR; Listwise: B-RankGPT; Pairwise: B-PRP, and our novel EcoRank.
\begin{figure*}[t]
    \centering
    \centering
    \begin{subfigure}[b]{0.5\textwidth}
      \centering
      \includegraphics[width=\linewidth]{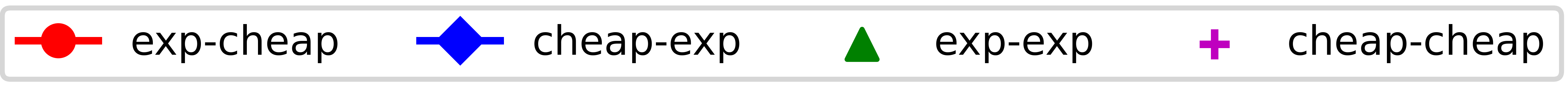}
    \end{subfigure}%
    
    \begin{subfigure}[b]{0.35\textwidth}
      \centering
      \includegraphics[width=\linewidth]{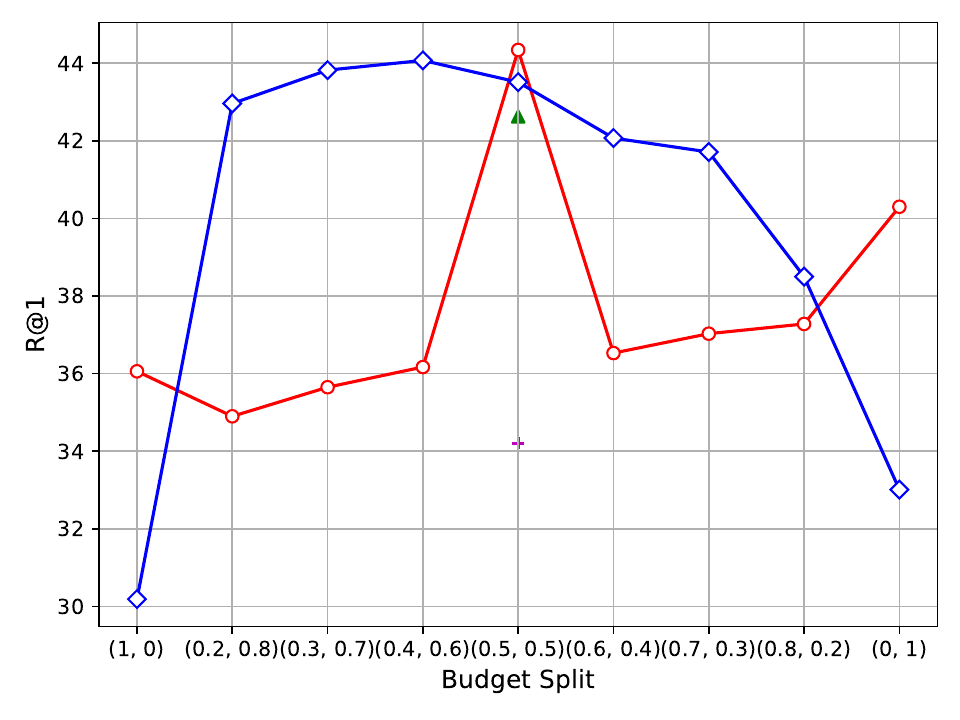}
      \caption{Dataset NQ}
      \label{fig:nq_graph}
    \end{subfigure}%
    ~
    \begin{subfigure}[b]{0.35\textwidth}
      \centering
      \includegraphics[width=\linewidth]{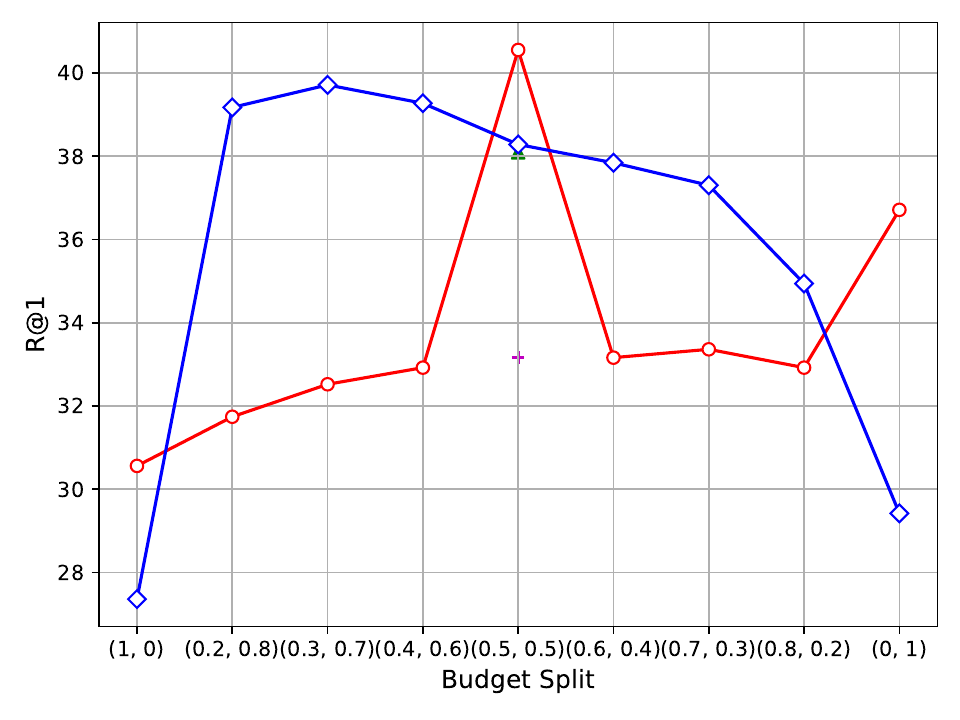}
      \caption{Dataset WQ}
      \label{fig:wq_graph}
    \end{subfigure}%
\caption{(a) and (b) subgraphs show the impact of our parameter choices in {\oursystem} for budget B2.}
    \label{fig:graph_figs}
\end{figure*}
\subsection{Main Results}
We show our main evaluation results for different budgets and ranking strategies in Table~\ref{tab:main_results} for $N=50$ passages. We choose the popular Mean Reciprocal Rank (MRR) and Recall@k as our evaluation metrics following previous work. All LLM-based approaches increase the initial ranking significantly except for B-UPR. 
%It shows that UPR is not a good choice for generation-only LLMs where we cannot get the scores of outputs. 
We see that, overall, for all budget categories, EcoRank surpasses all other approaches, even the supervised ones. For TREC DL 20 dataset, as there are many relevant passages given a query, B-PRP performs a little better for budget B1 and ours performs similarly.   
%Among the pointwise methods, Binary approach performs best. 
B-RankGPT is promising but it can only work with a high budget (i.e. B1) as GPT-3.5 is expensive. 
%Further we can also make the following observations.

\paragraph{Supervised vs Unsupervised.} We see that supervised models sometimes perform better than {\oursystem} in some datasets in B3. As we assume zero cost for supervised models, \textit{budget} is not directly applicable. They can process all the passages while budget-aware approaches can process only a few in B3. Even with this limitation, our methods perform better than supervised models in B1 and B2. Further, supervised models have the following restrictions: 1) training data may either not be available or be very difficult to collect, especially if the domain is niche, and 2) training data may incur high annotation costs. In real life systems, they are expected to perform worse. Unsupervised fine-tuning based approaches like InPars also do not perform as well as ours.

\paragraph{High to low budget analysis.} Among budget-aware methods, for higher budget (B1, B2), we see B-PRP method performing better than pointwise methods like binary classification in MRR and R@1 metrics. As we decrease our budget (B3), due to pairwise methods not processing enough passages, they perform worse than binary method in MRR. EcoRank achieves the best results in all budget categories, proving the efficiency of this approach. The gain of EcoRank with the second best approach increases from an average of 2\% for higher budget to 12\% for lower budget for R@1. This shows with lower budget constraint, our most efficient approach can perform really well.

% \paragraph{Optimizing R@1 vs R@10.} Pairwise methods focus more on R@1 as usually only one pass of the sorting algorithm can be performed with limited budget. Pointwise methods, as they are coarse-grained, focus more on R@k where $k > 1$. Thus we can see that pointwise methods score higher than pairwise in all budget categories in R@10. Likert pointwise approach has the best R@10 results in lower budget categories. There is an opportunity to optimize the evaluation metric based on user preference which we leave for future work.
\subsection{Analysis of our chosen parameters}
For {\oursystem} we have made some decisions regarding three sets of parameters: 1) The choices of prompting strategies, 2) The choices of expensive and cheap LLMs $\mathcal{L}_1$, $\mathcal{L}_2$, and 3) The choices of budget split between the prompts $x$ and $y$. We chose pointwise and pairwise strategies due to their \textit{cost-to-performance ratio} as seen in Table~\ref{tab:llm_variants}. We choose an expensive LLM for the first stage and a cheap LLM for the second stage with equal budget split $x=0.5$, $y=0.5$. The justification being that an expensive LLM is needed in the first stage to filter the important passages for the pairwise approach to focus on the second stage. As both stages play an equal part, an equal budget split is the appropriate choice.

We accompany our choices with an extensive evaluation performed on two datasets with all combinations of budget splits and LLM choices shown in Figure~\ref{fig:graph_figs}. The blue line shows the results if we use the cheaper LLM in the first stage. We see a declining performance than choosing an expensive LLM (shown in red) in stage one. As we move closer to an equal split, the performance keeps increasing till it reaches a peak and declines again as we move away. 
%Our equal budget split choice also brings the best results. 
For detailed results on chosen parameters, please refer to appendix~\ref{app:additional_results_splits}.

\subsection{Ablation studies of EcoRank}
In EcoRank, there are two main choices that impact the performance.

\paragraph{1) Hybrid prompt design.}We use a combination of pointwise and pairwise methods. If we consider the \textit{intermediate ranked list} (from Figure~\ref{fig:overview}) that is generated after the first stage as the final ranking, the performance decreases as only half budget is used. Further, if we only use one ranking prompt design (either Binary or B-PRP in Table~\ref{tab:main_results}) using full budget we also see a significant decrease in performance than EcoRank. Hybrid prompt design is crucial to get the maximum performance in a budget-constrained scenario. 

\paragraph{2) Cascading of LLMs.} We show the results of a variant of EcoRank where we do not cascade LLMs but keep the hybrid prompt design. We call this approach EcoRank-w/o-cascade. Only the expensive LLM is queried in both stages. Although an expensive LLM is more accurate, we see from Table~\ref{tab:main_results} that, this variant falls short of EcoRank. It still performs better than other methods. Using a cheaper LLM in the second stage can enable processing more passages resulting in better performance. Cascading of cheap and expensive LLMs is impactful to getting the maximum performance.

%% file: Tables/budget_category.tex
\begin{table}
\small
    \centering
    \begin{tabular}{c|c|c|c|c}
    \toprule
         Category &Cost&  T5-XL&  T5-L& GPT-3.5\\
         \midrule
         B1 &0.57c&  20000&  60000& 2000\\
         B2 &0.11c&  4000&  12000& 400\\
         B3 &0.05c&  2000&  6000& 200\\
         \bottomrule
    \end{tabular}
    \caption{Budget categories for different LLMs.}
    \label{tab:budget_category}
\end{table}

%% file: Tables/main_results.tex
\begin{table*}[t]
    \small
    \centering
    \begin{tabular}{l|l|l|l|l|l|l|l|l|l|l}
    \toprule
         \multirow{2}{*}{Method} &\multirow{2}{*}{Strategy}&  \multirow{2}{*}{LLM}&  \multicolumn{2}{c|}{NQ}&  \multicolumn{2}{c|}{WQ} & \multicolumn{2}{c|}{TREC DL 19}& \multicolumn{2}{c}{TREC DL 20}\\
          &&  &  \multicolumn{1}{c}{MRR}& \multicolumn{1}{c|}{R@1} & \multicolumn{1}{c}{MRR}&  \multicolumn{1}{c|}{R@1} & \multicolumn{1}{c}{MRR}& \multicolumn{1}{c|}{R@1}& \multicolumn{1}{c}{MRR}&\multicolumn{1}{c}{R@1}\\
         \midrule
         BM25 &-&  -&  32.49&  22.10  &  29.69&  18.89 & 48.15& 30.00& 69.67&56.86\\
         \midrule
 \multicolumn{11}{c}{\textbf{Supervised Models}}\\
 \midrule
 monot5-base& -& T5-base & 48.01&  39.11& 43.15& 33.21& \textbf{74.57}& \textbf{66.66}& 79.94&68.18\\
 monot5-3B& -& T5-3B & \textbf{50.47}& \textbf{41.66}&  \textbf{44.88}& \textbf{35.23}& 72.75& 63.33& 79.49&65.9\\
 monoBERT& -& BERT & 47.31& 38.25& 43.63&  33.8& 73.68& 66.66& \textbf{81.06}&\textbf{70.45}\\
 TART& -& T5-XL& 46.42& 37.47& 42.65& 33.07& 73.13& 60.00& 68.03&52.28\\
       \midrule
         \multicolumn{11}{c}{\textbf{Unsupervised: B1 - 0.57 cents per question}}\\
         \midrule
 InPars& -& T5-large& 42.73& 32.43& 41.49& 31.05& 59.63& 46.66& 74.72&61.36\\
         Binary &Point&  T5-XL&  46.77&  37.36&   42.09&  31.54 & 64.86& 50.00& 70.67&54.54\\
 Binary &Point& GPT-3.5& 36.10& 25.80& 34.06& 22.58& 53.76& 40.00& 59.99&45.45\\
         Likert &Point&  T5-XL&  39.49&  28.25&   39.81&  28.44 & 59.86& 43.33& 63.85&50.00\\
         B-UPR&Point&  T5-XL&  27.37&  17.10&   29.69&  18.89 & 26.17& 10.00& 36.12&25.00\\
         B-PRP&Pair&  T5-XL&  51.85&  45.04&   48.11&  41.14& 65.40& 56.66& 78.97&68.18\\
 B-PRP&Pair& GPT-3.5& 36.68& 29.00& 37.25& 30.01& 63.85& 53.33& 66.4&54.54\\
 B-RankGPT&List& GPT-3.5& 45.05& 37.83& 42.01& 34.10& 66.82& 56.66& 72.23&63.63\\
         {\oursystem}$^\dagger$&Hybr.&  T5-XL&  \textbf{52.76}&  \textbf{45.87}&   \textbf{48.94}&  \textbf{41.58} & \textbf{78.87}& \textbf{67.44}& \textbf{80.63}&\textbf{70.45}\\
        \midrule
 \multicolumn{11}{c}{\textbf{Unsupervised: B2 - 0.11 cents per question}}\\
 \midrule
 OpenAI Embedding& -& text-ada-002& 32.37& 22.10& 29.69& 18.89& 48.15& 30.00& 58.67&45.45\\
 InPars& -& T5-large& 43.92& 33.57& 42.11& 31.39& 60.39& 46.66& 72.59&54.54\\
 Binary &Point& T5-XL& 44.88& 36.06 & 40.84& 30.56& 64.66& 50.00& 70.67&54.54\\
 Likert &Point& T5-XL& 38.96& 28.00 & 39.17& 27.95& 59.86& 43.33& 63.78&50.00\\
 B-UPR&Point& T5-XL& 28.03& 17.45 & 29.69& 18.89& 25.62& 10.00& 35.96&25.00\\
 B-PRP&Pair& T5-XL& 45.97& 40.30 & 42.79& 36.71& 65.33& \textbf{56.66}& 74.44&59.09\\
 {\oursystem}-w/o-casc.&Hybr.& T5-XL& 48.56& 42.63 & 44.87& 38.09& 63.13& 50.00& 74.06&59.09\\
 {\oursystem} &Hybr.& T5-XL+L& \textbf{50.72}& \textbf{44.34} & \textbf{47.05}& \textbf{40.55}& \textbf{65.58}& 53.33& \textbf{80.22}&\textbf{70.45}\\
\midrule
 \multicolumn{11}{c}{\textbf{Unsupervised: B3 - 0.05 cents per question}}\\
 \midrule
 InPars& -& T5-large& 44.49& 34.79& 42.63& 32.72& 59.84& 46.66& 74.3&61.36\\
 Binary &Point& T5-XL& 43.06& 34.59 & 39.20& 29.42& 62.97& 50.00& 70.67&54.54\\
 Likert &Point& T5-XL& 38.31& 27.83 & 37.65& 26.91& 59.79& 43.33& 63.58&50.00\\
 B-UPR&Point& T5-XL& 29.53& 18.55 & 30.10& 19.43& 33.2& 16.66& 40.52&25.00\\
 B-PRP&Pair& T5-XL& 42.81& 36.98 & 39.45& 32.66& 64.02& 53.33& \textbf{78.93}   &68.18\\
 {\oursystem}-w/o-casc.&Hybr.& T5-XL& 44.13& 37.89 & 40.66& 33.85& 63.13& 50.00& 77.1&67.81\\
 {\oursystem} &Hybr.& T5-XL+L& \textbf{46.83}& \textbf{40.33} & \textbf{43.86}& \textbf{37.00}& \textbf{66.92}& \textbf{56.66}& 78.76&\textbf{70.45}\\
\bottomrule
    \end{tabular}
    \caption{Results (MRR and R@1) on all datasets for 50 passages. Best performing in all categories are marked bold. For B2 and B3, the budget is too low for B-RankGPT to have any impact hence it is omitted. $^\dagger$ For high budget B1, all 50 passages can be processed with T5-XL hence cascade of APIs is not needed in {\oursystem}.}
    \label{tab:main_results}
\end{table*}

%% file: Sections/f_related_work.tex
\section{Related Work} \label{sec:related}
To the best of our knowledge, we are the first to work on budget-constrained text re-ranking problem with LLMs. We divide the related work into LLMs in text re-ranking and Cost aware LLMs.

\paragraph{LLMs in text re-ranking.} There are three main zero-shot prompting strategies to re-rank initially ranked passages. They are pointwise~\citep{liang2022holistic, sachan2022improving, zhuang2023beyond}, listwise~\citep{sun2023chatgpt, ma2023zero, tang2023found}, and pairwise~\citep{qin2023large}, which we covered in details in Section~\ref{sec:ranking_algo}. Each strategy has their own strengths and may not work with all types of LLMs. Very recently another prompting strategy has been introduced called \textit{setwise}~\citep{zhuang2023setwise}, which is an improvement over listwise approach where instead of outputting an ordered list of documents, a single document which is the most relevant is given as output. Although this reduces the computational overhead of listwise and pairwise methods, it works best with LLMs which can output scores of generation. Other works like distillation~\citep{sun2023instruction}, RankVicuna~\citep{pradeep2023rankvicuna}, RankZephyr~\citep{pradeep2023rankzephyr} train an open source model with training data to improve listwise approaches. All these approaches do not consider the cost of LLMs and do not try to optimize the performance of text rankers with a budget constrain.
Prior to recent efforts with LLMs in text re-ranking, most works focused on the supervised ranking problem using monoT5~\citep{nogueira2020document} or BERT~\citep{zhuang2021ensemble} where they trained a pre-trained LM (PLM) for re-ranking tasks. Other supervised methods focus on generating data to train PLMs like InPars~\citep{bonifacio2022inpars}, Promptagator~\citep{dai2022promptagator}, ExaRanker~\citep{ferraretto2023exaranker}, SPTAR~\citep{peng2023soft}, HyDE~\citep{gao2022precise} etc. They mainly use LLMs as an auxiliary tool to support the training of PLMs and thus different from the scope of this paper.

\paragraph{Cost-aware LLMs.} There are some works which focus on cost-aware applications of LLMs but in other areas than text re-ranking. FrugalGPT~\citep{chen2023frugalgpt} uses LLM cascading to reduce the cost of API calls but they require a trained model to score the generation quality similar to FORC~\citep{vsakota2023fly} which uses a trained meta-model to predict performance of LLMs. These trained models require fine-tuning data which may be difficult to obtain. MoT~\citep{yue2023large} uses answer sampling strategy which is not applicable in text re-ranking as LLMs output fixed tokens instead of open-ended. Other works focus on optimizing API calls by using a neural caching system with a student model~\citep{ramirez2023cache}. None of these apply to our problem statement as we aim to optimize the performance in text re-ranking in a fully unsupervised fashion.

%% file: Sections/h_conclusion.tex
\section{Conclusion}\label{sec:conclusion}
We contribute to the first efforts of budget-constrained text re-ranking with LLMs and have identified that existing works fail to consider budget while optimizing performance in text re-ranking. We propose a suite of budget-constrained methods with various ranking prompt designs and LLMs and extensively evaluate them on four datasets. Our most efficient method EcoRank, which is a two-layered pipeline that optimizes vast spaces of decisions, achieves a gain of 14\% on MRR and R@1 than other approaches. 

%% file: Sections/i_limitation.tex
The first limitation is that we do not consider all possible LLM APIs or open source models. We pick a representative subset of them.
Also, given the combinatorial cost, we consider limited combinations of re-ranking prompts in building EcoRank. We employ pointwise and pairwise, as they perform best in cost-to-performance ratio. Some of our design choices are discrete (i.e. choosing the LLMs in a static way). We also contribute to the first efforts towards automating our pipeline. The experiments are provided in Appendix~\ref{app:automate}.
Further, we experiment with 50 passages per query and thus set lower budget categories due to the high cost of conducting the experiments. Ideally, more passages can be considered (i.e. 1000) with budget limits increasing proportionally. However, this choice does not impact the efficacy of our budget-constrained solutions.
%A limitation of our budget-constrained approaches is the adaptation LLMs. We use a static choice of using Flan T5-XL as the expensive LLM and Flan T5-large as the cheap LLM as they perform the best among the LLMs that are available till date. If any LLM is introduced in the future which has better performance in text re-ranking and is cost-effective, that maybe a better choice than using Flan T5. Similarly, if the prices of other LLM APIs that are available now but we are not considering (i.e. GPT 3.5) drop significantly, our choices of LLMs may not be the best performing.

Another limitation is about the optimization of evaluation metrics. We optimize for R@1 as in QA task we typically care about the top few re-ranked passages. We see some prompting strategies (i.e. likert) performing best for R@10 but not R@1. Users may have preference to optimize for R@10 but we do not take that into consideration and leave for future work.

%% file: Sections/g_appendix.tex
\section{Pointwise Methods Ranking}
\label{app:rank_pointwise}
\paragraph{Binary Classification.} We ask the LLM to predict \textit{Yes} or \textit{No} given each passage from the initial list whether it is relevant to the query $q$. 
%The prompt is shown in Figure~\ref{fig:prompts}. 
Multiple passages may get the output of "Yes" as the passages are initially retrieved by a retriever like BM25. For a passage, if the output is not parseable or we exhaust the budget, we fall back to the initial ranking. This results in a list of re-ranked passages where all the "Yes" passages are grouped together at the top sorted by the initial ordering, unprocessed passages at the middle, and "No" passages at the bottom. More passages can be processed with LLMs as they require less number of tokens as input and output than other methods.

\paragraph{Likert Classification.} We ask the LLM to classify a passage into either of the following three groups: \textit{Very related, Somewhat related} or \textit{Unrelated}. 
This is an attempt to fine-grain the passages a little bit more with the same number of tokens. 
However, it may be difficult for an LLM to differentiate between \textit{Very related} and \textit{Somewhat related}. Similar to the previous method, we group the \textit{Very related} passages at the top, then \textit{Somewhat related}, then unprocessed passages, and \textit{Unrelated} passages at the bottom.

\input{Tables/cost_compare}
\section{Cost Estimation of Baselines}
\label{app:supervised_cost}
Our goal is to find the best performing system in passage re-ranking in a budget-constrained scenario. We define budget as a financial unit, where we estimate the prices for different paid APIs. For supervised models in our baseline, it becomes extremely challenging to estimate the budget in terms of financial units as the models are hosted offline. They are not typically hosted by organizations where users can pay for the service. Thus, the costs between supervised and unsupervised methods are not directly comparable. For simplicity, in this work, we assume zero cost for supervised models. However, there are other costs exclusive to supervised models that are not negligible. These costs are likely to worsen the performance of these systems in real life. For example:
    \paragraph{1) Data collection and annotation costs.} It is difficult to find good-quality training data for any task. It becomes especially challenging for niche domains where there are not enough training data. The time to find such data can be costly. Further, the training data needs gold labels. Annotating labels can be very expensive. For example, it may cost up to 10 USD to annotate 50 data points using Amazon MTurk~\citep{mturk}.
    \paragraph{2) Training time costs.} Supervised models are trained with huge amounts of data. The training time can be very costly. For example, monoT5 takes 30 hours~\citep{nogueira2020document} while monoBERT takes 160 hours~\citep{nogueira2019passage}. For unsupervised non-LLM based approach InPars, it also has some amount of training time.
    \paragraph{3) Computing resources.} Supervised models need high computing resources such as GPUs or TPUs. Although they are considered a one-time cost, they can be significant for a small to mid-size organization. InPars also needs computing resources to host the fine-tuned models.

We summarize and compare the different costs for supervised and unsupervised systems with {\oursystem} in Table~\ref{tab:cost_compare}.

\section{Additional Evaluation}
\label{app:additional_results}
\subsection{Results on various budget splits and LLM choices}\label{app:additional_results_splits}
We present various combinations of budget splits and LLM choices on the two stages of EcoRank in Table~\ref{tab:splits}. We see that, our design choice of an equal budget split with expensive LLM on the first stage and cheaper LLM on the second stage provides the best results.
\input{Tables/splits}

\subsection{{\oursystem} with $n$-stages}
\label{app:n_stages}
\input{Tables/n_stages}
In this subsection, we show the results of {\oursystem} with three and four stages in Table~\ref{tab:n_stages}. We choose an equal budget split as this has been shown to perform the best in Table~\ref{tab:splits}. We consider the combinations of \textit{pointwise} and \textit{pairwise} strategies as \textit{listwise} strategy has been shown to provide worse results in Table~\ref{tab:llm_variants}. Please also note the following constraints on a pipeline with more than two stages:
\begin{itemize}
    \item Consecutive stages should not have the same prompting strategy. For example, in \textit{pointwise -> pointwise -> pairwise}, as the passages have already been processed once, using the same strategy again will not be productive.
    \item Subsequent strategies should only use a more expensive LLM. Applying a cheaper LLM on a prompting strategy after applying an expensive LLM on the same strategy will deteriorate the performance. For example, \textit{pointwise(exp) -> pairwise(cheap) -> pointwise(cheap)}.
\end{itemize}
Keeping these constraints, we experiment with three and four-stage pipelines. Due to exponentially high computational and resource costs, we do not show all permutations of budgets and LLM choices.
\paragraph{Three stages.} For the three stages, we choose pointwise -> pairwise -> pointwise strategy. For LLM choices, we consider the first stage to use cheap LLM and show combinations in the second and third. We see a diminishing return as we add more stages.
\paragraph{Four stages.} We choose pointwise -> pairwise -> pointwise -> pairwise strategy. For LLM choices, we use different combinations. Here we also see that the performance is not as good as {\oursystem} with two stages. Adding extra stages to our pipeline incurs less budget in each individual stage, resulting in only the top few passages being processed over and over again.

\subsection{R@10 results}
\label{app:r10_results}
In QA~\citep{rashid2021quax,meem2024pat} or text re-ranking pipelines, typically only the top few passages are considered. In most cases, we only need one good passage to answer a query, and hence R@1 is the most important. Nevertheless, we also show R@10 results for completeness in Table~\ref{tab:r10_results}.
\paragraph{Optimizing R@1 vs R@10.} Among the unsupervised LLM-based approaches, pairwise methods focus more on R@1 as usually only one pass of the sorting algorithm can be performed with a limited budget. Pointwise methods, as they are coarse-grained, focus more on R@k where $k > 1$. Thus we can see that pointwise methods score higher than pairwise in all budget categories in R@10. Overall, InPars, which is an unsupervised fine-tuned model, performs best in R@10 for lower budget categories. It is mostly because it can process all of the passages as inference does not cost anything.
\input{Tables/automate}
\input{Tables/r10}

\subsection{Selecting budget categories}
\label{app:selecting_budget}
For our evaluation, we have selected three budget categories, a higher budget (0.57c) so that most of the passages (among 50 total) can be processed, a medium budget (0.11c), and a low budget(0.05c) so that only a few passages can be processed. These are the steps that we took to come up with the budget and number of tokens:
\begin{itemize}
    \item First we chose the best performing LLM from Table~\ref{tab:llm_variants}, which is Flan T5-XL.
    \item We calculated the total number of tokens required to process $\sim$80\% of the passages using Flan T5-XL using pointwise method to come up with a high budget. We determined this value to be 20000 for T5-XL.
    \item We came up with the medium budget B2 and B3 by dividing the high budget (20000) by half. Low-budget B3 can only process $\sim$20\% of the passages.
    \item For other LLMs such as T5-L, GPT-3.5, etc., we determined the token values using the cost estimate from Table~\ref{tab:llm_variants}. For example, If Flan T5-XL can process 20000 tokens, then GPT-3.5 can process 2000 tokens as it is 10 times more expensive. These cost estimates are taken from the pricing web pages.
    \item Finally, we converted the number of tokens into USD using the pricing webpages of LLM API services such as OpenAI, TextSynth. We show these numbers in Table~\ref{tab:budget_category}.
\end{itemize}

\subsection{Results on LLMs with lower cost}
\label{app:additional_results_weaker}
Table~\ref{tab:appendix_weaker_results} shows evaluation results of different strategies with different budgets on a cheaper LLM. For a cheaper LLM we consider T5-large which is approximately 3x cheaper than T5-XL. However, as we saw with GPT vs T5-XL where T5-XL was the cheaper LLM, still it produced better results, this is not the case here. T5-L produces worse results than T5-XL but still comparable as they are trained in a similar fashion. We see that EcoRank has the best results than other budget-constrained approaches.

\begin{table*}[t]
    \small
    \centering
    \begin{tabular}{l|l|l|l|l|l|l|l|l}
   \toprule
         \multirow{2}{*}{Method} &\multirow{2}{*}{Strategy}&  \multirow{2}{*}{LLM}&  \multicolumn{3}{c|}{NQ}&  \multicolumn{3}{c}{WQ}\\
          &&  &  \multicolumn{1}{c}{MRR}& \multicolumn{1}{c}{R@1}& \multicolumn{1}{c|}{R@10} & \multicolumn{1}{c}{MRR}&  \multicolumn{1}{c}{R@1}& \multicolumn{1}{c}{R@10}\\
         \midrule
         BM25 &-&  -&  32.49&  22.10&  54.45&  29.69&  18.89& 52.16\\
         \midrule
         \multicolumn{8}{c}{\textbf{B2 - 0.11 cents per question}} &\\
         \midrule
         Binary &Pointwise&  T5-large&  41.39&  30.19&  63.65&  39.00&  27.36& 61.90\\
         Likert &Pointwise&  T5-large&  40.20&  29.22&  61.66&  38.81&  27.70& 61.17\\
         B-UPR&Pointwise&  T5-large&  26.01&  15.90&  48.75&  29.55&  19.09& 52.01\\
         B-PRP&Pairwise&  T5-large&  40.96&  33.01&  58.03&  37.66&  29.42& 55.06\\
         {\oursystem}-w/o-cascade &Hybrid&  T5-large&  \textbf{46.08}&  \textbf{36.56}&  \textbf{64.48}&  \textbf{43.22}&  \textbf{33.16}& \textbf{62.79}\\
         \midrule
 \multicolumn{8}{c}{\textbf{B3 - 0.05 cents per question}} &\\
 \midrule
 Binary &Pointwise& T5-large& 41.21& 30.08& \textbf{63.60}& 39.00& 27.36&\textbf{61.90}\\
 Likert &Pointwise& T5-large& 39.99& 29.08& 61.46& 38.71& 27.41&61.81\\
 B-UPR&Pointwise& T5-large& 26.01& 15.87& 48.78& 29.65& 19.14&52.31\\
 B-PRP&Pairwise& T5-large& 42.10& 33.82& 58.94& 38.18& 28.54&56.25\\
 {\oursystem}-w/o-cascade &Hybrid& T5-large& \textbf{44.57}& \textbf{35.62}& 61.08& \textbf{41.63}& \textbf{31.84}&60.28\\
 \midrule
 \multicolumn{8}{c}{\textbf{B4 - 0.025 cents per question}} &\\
 \midrule
 Binary &Pointwise& T5-large& 40.38& 29.83& \textbf{60.30}& 38.08& 26.91&\textbf{59.84}\\
 Likert &Pointwise& T5-large& 39.5& 28.91& 59.14& 37.68& 27.06&58.02\\
 B-UPR&Pointwise& T5-large& 32.37& 22.10& 54.45& 29.69& 18.89&52.16\\
 B-PRP&Pairwise& T5-large& 38.77& 30.83& 55.42& 35.30& 26.72&53.05\\
 {\oursystem}-w/o-cascade &Hybrid& T5-large& \textbf{41.87}& \textbf{32.90}& 58.06& \textbf{39.57}& \textbf{30.36}&56.69\\
 \bottomrule
    \end{tabular}
    \caption{Results (MRR and R@K) on NQ and WQ datasets for T5-large. Best performing are marked bold.}
    \label{tab:appendix_weaker_results}
\end{table*}
\section{Automating Discrete Choices}
\label{app:automate}
In {\oursystem}, we made some static decisions like using an expensive LLM in the first stage, choosing pointwise in the first stage etc. As first efforts towards automating the pipeline, we pursued two ideas and performed experiments to automate the \textbf{choosing of LLMs} decision. Below we describe the setup and results.

\paragraph{Approach 1: Assess question difficulty through pointwise.} The idea is to assess the difficulty of the question. The intuition being, that an expensive LLM is needed if the question is difficult otherwise we use a cheaper LLM if it is a simpler question. To assess the question's difficulty, we first use pointwise promting strategy to process the top-$m$ passages. If the number of \textit{"Yes"} generated by the LLM is greater than a threshold $t$, we classify the question to be difficult. Once a question has been classified as difficult, we will continue using an expensive LLM for both stages and not use a cheaper LLM at all. If it is not a difficult question, we will switch to a cheaper LLM and keep using it for the rest of the passages for both stages. We call this approach \texttt{EcoRank-Auto1}. For our experiment, we use $m=8$ and $t=4$.

\paragraph{Approach 2: Assess question difficulty through BM25 scores.} The intuition is, that for a difficult question, the standard deviation of the BM25 scores of the passages will be greater than a not-difficult question. If a question is difficult, the scores of the passages retrieved by BM25 will be very close, resulting in smaller stdev. We call this approach \texttt{EcoRank-Auto2}. We set a threshold $st=1.5$, if it is less than this the question is difficult.

We set the parameters manually as these are initial experiments and the goal is to see the potential of the ideas. A validation set can be used to fix the parameters.
\paragraph{Results.} We show the results in Table~\ref{tab:automate}. While the approaches did not surpass {\oursystem}, it still performs better than other budget-aware methods like B-PRP, Binary, etc. We understand this is not a fully automated system. These experiments show that there is good potential for us and other researchers to automate {\oursystem} in the future.

%% file: Tables/cost_compare.tex
\begin{table*}[t]
\small
    \centering
    \begin{tabular}{c|c|c|c|c|c}
    \toprule
         Method&Approach&  Data Costs&  Training Time Costs& Computing Costs &Inference Costs\\
         \midrule
         monoT5&Supervised&  High&  30 Hours& \cmark &0\\
         monoBERT&Supervised&  High&  160 Hours& \cmark &0\\
         InPars&Unsupervised&  Low&  5 Mins& \cmark &0\\
         \midrule
 {\oursystem}& Unsupervised& 0& 0& 0& \cmark\\
 \bottomrule
    \end{tabular}
    \caption{Cost comparison of different methods.}
    \label{tab:cost_compare}
\end{table*}

%% file: Tables/splits.tex
\begin{table*}[t]
    \small
    \centering
    \begin{tabular}{l|l|l|l|l|l|l|l}
    \hline
         \multirow{2}{*}{Budget Splits (x,y)}&LLM Choice&  \multicolumn{3}{c|}{NQ}&  \multicolumn{3}{c}{WQ}\\
          &First stage - Second stage&  \multicolumn{1}{c}{MRR}& \multicolumn{1}{c}{R@1}& \multicolumn{1}{c|}{R@10} & \multicolumn{1}{c}{MRR}&  \multicolumn{1}{c}{R@1}& \multicolumn{1}{c}{R@10}\\
         \hline
         \multirow{2}{*}{1, 0}&exp, \_&  44.88&  36.06&  60.72&  40.84&  30.56& 59.64\\
         &cheap, \_&  41.39&  30.19&  \textbf{63.65}&  39.00&  27.36& 61.90\\
         \hline
 \multirow{2}{*}{0.2, 0.8}&exp, cheap& 43.56& 34.90& 58.44& 40.45& 31.74&57.08\\
         &cheap, exp&  49.42&  42.96&  61.38&  46.09&  39.17& 60.08\\
         \hline
         \multirow{2}{*}{0.3, 0.7}&exp, cheap&  44.26&  35.65&  58.58&  41.32&  32.52& 57.67\\
         &cheap, exp&  50.30&  43.52&  61.57&  46.67&  39.71& 61.12\\
 \hline
 \multirow{2}{*}{0.4, 0.6}&exp, cheap& 44.66& 36.17& 59.35& 41.85& 32.92&58.02\\
 &cheap, exp& 50.62& 44.07& 62.13& 46.50& 39.27&61.36\\
         \hline
         \multirow{2}{*}{0.5, 0.5}&exp, cheap&  \textbf{50.72}&  \textbf{44.34}&  62.46&  \textbf{47.05}&  \textbf{40.55}& 60.33\\
 &cheap, exp& 50.37& 43.51& 63.60& 46.02& 38.28&62.05\\
 \hline
 \multirow{2}{*}{0.6, 0.4}&exp, cheap& 45.34& 36.53& 60.02& 42.22& 33.16&58.36\\
 &cheap, exp& 49.19& 42.07& \textbf{63.65}& 45.56& 37.84&\textbf{61.86}\\
 \hline
 \multirow{2}{*}{0.7, 0.3}&exp, cheap& 45.67& 37.03& 60.08& 42.50& 33.36&59.10\\
 &cheap, exp& 48.76& 41.71& \textbf{63.65}& 45.07& 37.30&\textbf{61.86}\\
 \hline
 \multirow{2}{*}{0.8, 0.2}&exp, cheap& 45.81& 37.28& 59.52& 42.42& 32.92&59.30\\
 &cheap, exp& 46.10& 38.50& \textbf{63.65}& 43.30& 34.94&\textbf{61.86}\\
 \hline
 \multirow{2}{*}{0, 1}&\_, cheap& 40.96& 33.01& 58.03& 37.66& 29.42&55.06\\
 &\_, exp& 45.87& 40.30& 57.72& 42.79& 36.71&55.65\\
 \hline
    \end{tabular}
    \caption{Results of various budget splits and LLM choices on the stages of EcoRank for budget B2.}
    \label{tab:splits}
\end{table*}

%% file: Tables/n_stages.tex
\begin{table*}[t]
    \small
    \centering
    \begin{tabular}{l|l|l|l|l|l|l|l|l}
    \toprule
         \multirow{2}{*}{Method} &\multirow{2}{*}{Num. Stages}&  \multirow{2}{*}{LLM Choice}&  \multicolumn{3}{c|}{NQ}&  \multicolumn{3}{c}{WQ}\\
          &&  &  \multicolumn{1}{c}{MRR}& \multicolumn{1}{c}{R@1}& \multicolumn{1}{c|}{R@10} & \multicolumn{1}{c}{MRR}&  \multicolumn{1}{c}{R@1}& \multicolumn{1}{c}{R@10}\\
         \midrule
         {\oursystem}&2&  exp, cheap&  \textbf{50.72}&  \textbf{44.34}&  62.46&  \textbf{47.05}&  \textbf{40.55}& 60.33\\
 {\oursystem}& 2& cheap, exp& 50.37& 43.51& \textbf{63.60}& 46.02& 38.28&\textbf{62.05}\\
         \midrule
         {\oursystem}&3&  cheap, cheap, exp&  47.51&  39.88&  60.08&  42.95&  33.9& 59.05\\
 {\oursystem}&3& cheap, exp, exp& 47.87& 41.02& 60.41& 44.02& 36.41&59.25\\
    \midrule
         {\oursystem}&4&  cheap, cheap, exp, exp&  46.06&  38.06&  59.72&  42.72&  33.85& 58.21\\
         {\oursystem}&4&  exp, cheap, cheap, exp&  49.1&  43.7&  61.82&  46.96&  39.96& 60.58\\
        \bottomrule
    \end{tabular}
    \caption{Results of {\oursystem} with 3 and 4 stages on NQ and WQ datasets for budget B2 for equal budget split.}
    \label{tab:n_stages}
\end{table*}

%% file: Tables/automate.tex
\begin{table}[t]
    \small
    \centering
    \begin{tabular}{l|l|l}
    \toprule
         \multirow{2}{*}{Method} &  \multicolumn{2}{c}{NQ}\\
          &  \multicolumn{1}{c}{MRR}& \multicolumn{1}{c}{R@1} \\
         \midrule
         {\oursystem}&  \textbf{46.83}&  \textbf{40.33}\\
    \midrule
         {\oursystem} - Auto1&  43.71&  36.42\\
         {\oursystem} - Auto2&  44.56&  37.06\\
        \bottomrule
    \end{tabular}
    \caption{Results of  automated LLM choosing approaches for NQ dataset for budget B3 with equal budget split.}
    \label{tab:automate}
\end{table}

%% file: Tables/r10.tex
\begin{table*}[t]
    \small
    \centering
    \begin{tabular}{l|l|l|l|l|l|l}
    \toprule
         \multirow{2}{*}{Method} &\multirow{2}{*}{Strategy}&  \multirow{2}{*}{LLM}&  NQ&  WQ& DL 19& DL 20\\
          &&  &  \multicolumn{1}{c|}{R@10} & \multicolumn{1}{c|}{R@10} & \multicolumn{1}{c|}{R@10}& \multicolumn{1}{c}{R@10}\\
         \midrule
         BM25 &-&  -&  54.45&  52.16& 93.33& 92.15\\
         \midrule
 \multicolumn{7}{c}{\textbf{Supervised Models}}\\
 \midrule
 monot5-base& -& T5-base & 64.65& 62.94& 96.66& 100.0\\
 monot5-3B& -& T5-3B & \textbf{66.59}&  \textbf{63.97}& \textbf{100.0}& \textbf{100.0}\\
 monoBERT& -& BERT & 65.18& 62.94& 96.66& 100.0\\
 TART& -& T5-XL& 64.37& 62.64& 100.0&97.72\\
 \midrule
 \multicolumn{7}{c}{\textbf{B1 (unsup)}}\\
         \midrule
 InPars& -& T5-large& 62.40& 61.95& 86.66& \textbf{97.72}\\
         Binary &Point&  T5-XL&  64.37&   61.76& 90.00& 95.45\\
 Binary &Point& GPT-3.5& 58.19& 56.74& 96.66& 90.9\\
         Likert &Point&  T5-XL&  63.07&   62.05& 93.33& 93.18\\
         B-UPR&Point&  T5-XL&  48.45&   52.16& 63.33& 59.09\\
         B-PRP&Pair&  T5-XL&  63.71&   60.53& \textbf{96.66}& 95.45\\
 B-PRP&Pair& GPT-3.5& 52.00& 52.16& 93.33& 86.36\\
 B-RankGPT&List& GPT-3.5& 58.33& 56.88& 90.00& 90.9\\
         {\oursystem}&Hybr.&  T5-XL&  \textbf{65.45}&   \textbf{63.58}& 95.34& 95.45\\
         \midrule
 \multicolumn{7}{c}{\textbf{B2 (unsup)}}\\
 \midrule
 InPars& -& T5-large& 63.51& \textbf{61.66}& 83.33& \textbf{100.0}\\
 Binary &Point& T5-XL& 60.72& 59.64& 86.66& 95.45\\
 Likert &Point& T5-XL& 61.57& 60.92& 93.33& 93.18\\
 B-UPR&Point& T5-XL& 49.55& 52.16& 60.00& 61.36\\
 B-PRP&Pair& T5-XL& 57.72& 55.65& \textbf{96.66}& 95.45\\
 {\oursystem}-w/o-casc.&Hybr.& T5-XL& 58.31& 56.79& 83.33& 95.45\\
 {\oursystem} &Hybr.& T5-XL+L& \textbf{62.46}& 60.33& 86.66& 95.45\\
 \midrule
 \multicolumn{7}{c}{\textbf{B3 (unsup)}}\\
 \midrule
 InPars& -& T5-large& \textbf{63.24}& \textbf{62.25}& 83.33& \textbf{97.72}\\
 Binary &Point& T5-XL& 58.31& 57.08& 83.33& 95.45\\
 Likert &Point& T5-XL& 59.11& 57.77& 93.33& 90.9\\
 B-UPR&Point& T5-XL& 53.51& 54.80& 83.33& 77.27\\
 B-PRP&Pair& T5-XL& 54.45& 53.65& \textbf{96.66}& 95.45\\
 {\oursystem}-w/o-casc.&Hybr.& T5-XL& 56.28& 55.21& 86.66& 90.9\\
 {\oursystem} &Hybr.& T5-XL+L& 57.72& 56.69& 86.66& 90.9\\
\bottomrule
    \end{tabular}
    \caption{Results (R@10) for all datasets.}
    \label{tab:r10_results}
\end{table*}